# Application of S-Transform on Hyper kurtosis based Modified Duo Histogram Equalized DIC images for Pre-cancer Detection

Sabyasachi Mukhopadhyay[1], Soham Mandal[2], Sawon Pratiher[3], Ritwik Barman[1], M. Venkatesh[3], Nirmalya Ghosh[1], Prasanta K Panigrahi[1]

[1]Indian Institute of Science Education and Research Kolkata, India
[2]Institute of Engineering & Management Kolkata, India
[3]Indian Institute of Technology Kanpur, India

*Abstract*—Our proposed hyper kurtosis based histogram equalized DIC images enhances the contrast by preserving the brightness. The evolution and development of precancerous activity among tissues are studied through S-transform (ST). The significant variations of amplitude spectra can be observed due to increased medium roughness from normal tissue were observed in time-frequency domain. The randomness and inhomogeneity of the tissue structures among human normal and different grades of DIC tissues is recognized by ST based time-frequency analysis. This study offers a simpler and better way to recognize the substantial changes among different stages of DIC tissues, which are reflected by spatial information containing within the inhomogeneity structures of different types of tissue.

## I. INTRODUCTION

We know that the spatial fluctuation of refractive index in biological tissues exhibits self-similar behavior containing morphological information which can be exploited for disease detection purpose. In our previous work, we used multifractal detrended fluctuation analysis (MFDFA), wavelet based approach and wavelet based MFDFA approach for pre-cancer detection purpose [1-3].

Time-frequency distribution (TFD) provides much useful information in the analysis of the non-stationary signals. The basic idea of this time-frequency representation is to quantify the changes in the frequency content of signals with time. In analyzing biological signals, time-frequency has been found to be a powerful tool [4]. Most of the TFD techniques are based on Wigner-Ville distribution (WVD). The 2-D smoothed versions of the WVD are referred as Quadratic TFDs (QTFD). As an example of QTFD, STFT (Short Time Fourier Transform), WT (Wavelet Transform) can be mentioned. A recently developed tool in the QTFD family is S-transform (ST) which is a hybrid of the CWT (Continuous Wavelet Transform) and STFT. ST uses a Gaussian window with a variable analyzing length. It preserves the phase information by using a Fourier kernel in the signal decomposition [5]. The low resolution of the STFT and the absence of phase information in the CWT are overcome by ST. ST, not only preserve the phase information, but also provides a better time-frequency resolution. Hence S-transform based approach is well-suited for phase synchrony analysis of a wide class of signals [10]. In this paper, S-transform has been applied on hyper kurtosis based duo histogram equalized DIC images. Amplitude spectra of S-transform clearly points out the differences of healthy and precancerous tissues in time frequency domain.

## II. THEORETICAL BACKGROUND

### A. Hyper Kurtosis based Modified Duo Histogram Equalization (HKMDHE)

There are several histogram equalization methods like CLAHE [12], Bi histogram equalization [7]. Our proposed hyper kurtosis based modified duo histogram equalization (HKMDHE) method has been explained below. The modified mean can be obtained as:

$$MM = \sqrt{m \pm \beta}, \beta = \frac{E(X-m)^6}{\sigma} \quad \text{........(1)}$$

Here 'mean' is the average pixel value of an image. The Positive sign is given when kurtosis is negative otherwise it is negative. $\sigma$ is the standard deviation of the distribution of $X$ which denotes the intensity.

### B. Peak Signal to Noise Ratio (PSNR)

2.2 For an input image $X(i,j)$ of size $M \times N$, the output image $Y(i,j)$ of same size will produce PSNR value as:

$$PSNR(dB) = 20\log_{10}\left(\frac{\max(Y(i,j))}{RMSE}\right) \quad \text{........(2)}$$

Here $i = 1,....,M$ and $j = 1,....,N$.
The RMSE (Root Mean Square Error) can be calculated

as: $\left(\dfrac{\sum_{i=1}^{M}\sum_{j=1}^{N}(X(i,j)-Y(i,j))^2}{M \times N}\right)^{\frac{1}{2}} \quad \text{........(3)}$

### C. Absolute Modified Mean Brightness Error (AMMBE)

The value of AMMBE can be calculated as:

$$AMMBE(X,Y) = |X_{MM} - Y_{MM}| \quad \text{......................(4)}$$

*D. S-Transform*

In ST, a given time series $x(t)$ is multiplied with a Gaussian window located at $t = \tau$ followed by taking a fourier transform of their product. So, the definition as follows [5]:

$$s(t,f,\sigma) = \int_{-\infty}^{+\infty} x(\tau) g(t-\tau,\sigma) e^{-j2\pi f\tau} d\tau \quad \text{..................(5)}$$

ST uses a scaled Gaussian window $g(t-\tau,\sigma)$ with midpoint at $\tau = t$. Due to the scaled contraction of $g(t-\tau,\sigma)$, the performance of ST becomes more localized around $t$ when $f$ increases. Here,

$$g(t-\tau,\sigma) = \frac{1}{\sigma\sqrt{2\pi}} e^{\frac{-(t-\tau)^2}{2\sigma^2}} \quad \text{......................(6)}$$

Now selecting $\sigma = \frac{1}{f}$, equation (2) becomes,

$$g\left(t-\tau,\frac{1}{f}\right) = \frac{f}{\sqrt{2\pi}} e^{\frac{-(t-\tau)^2 f^2}{2}} \quad \text{......................(7)}$$

This step is done for providing advantage over STFT. This is why ST is defined as Adaptive STFT. Even ST is also defined as a CWT multiplied by a phase factor. It can be written as:

$$s(t,f) = e^{-j2\pi ft} WT(t,d) \quad \text{......................(8)}$$

Where $d$ is a scale factor inversely proportional to $f$ i.e.,

$$WT(t,d) = \int_{-\infty}^{+\infty} x(\tau) g(t-\tau,\sigma) d\tau \quad \text{......................(9)}$$

Now using (3) and (4), it can be written as:

$$g\left(t-\tau,\frac{1}{f}\right) = \frac{f}{\sqrt{2\pi}} e^{\frac{-(t-\tau)^2 f^2}{2}} e^{-j2\pi f(t-\tau)} \quad \text{..................(10)}$$

In ST, the oscillatory exponential kernel $e^{-j2\pi f(t-\tau)}$ localizes the frequency referred as frequency modulated component which is a stationary part. It also localizes the real and imaginary parts of the phase spectrum. The slowly varying Gaussian envelope localizes the amplitude modulation which is a translated part. This is another advantage of ST over CWT. Amplitude of the CWT is large for lower frequencies, while at higher frequencies amplitude diminishes.

### III. METHODOLOGY

DIC images are at first processed through HKMDHE method in order to get optimum contrast followed by unfolding of data. This unfolded data are processed with S-transform. This histogram equalized processed has been performed such that signal processing tools like S-transform can more prominently classify the normal and disease tissues.

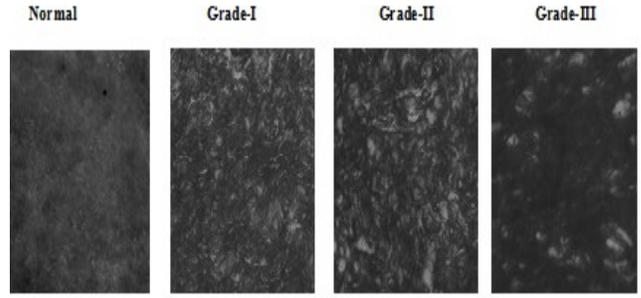

Fig-1 DIC images of Stromal region from Normal to Grade-III tissues respectively

### IV. RESULTS & DISCUSSIONS

DIC images contain the information of medium refractive index fluctuation. To extract this information properly, we need to remove the extra perturbations like speckle noise which is inherent to the DIC images as Laser sources are used. Before that we need to make the contrast enhancement by preserving the brightness as biomedical images are of low contrast images [11]. Contrast enhancement will help us to extract more prominent features in DIC images. The figures after applying MDHE over DIC images of stromal regions are shown in fig 2.

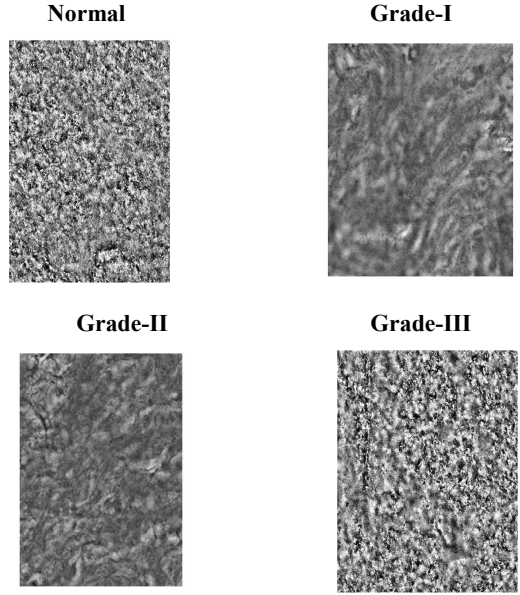

Fig-2 DIC images of Stromal region after applying HKMDHE from Normal to Grade-III tissues respectively

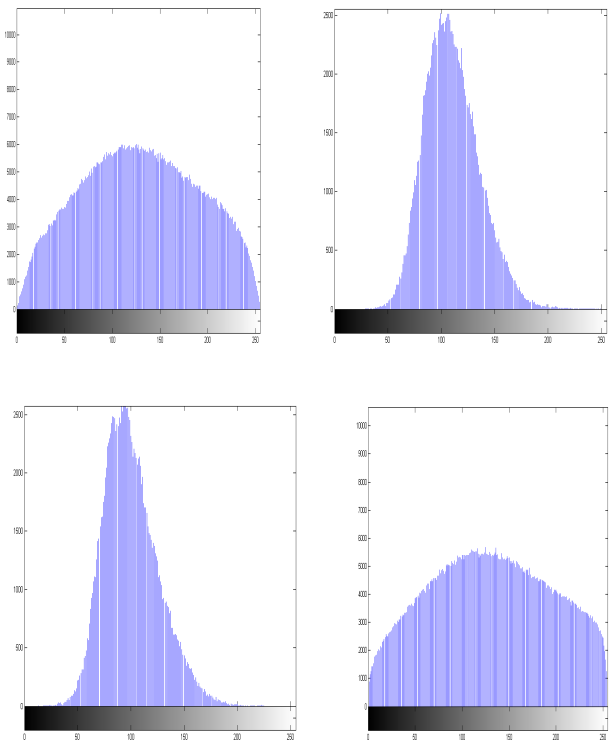

Fig.3 – MDHE of Corresponding Normal, Grade-I, Grade-II and Grade-III Images respectively. (X Axis: Pixel Intensity Values [0-255] & Y Axis: Number of Pixels)

The RMSE, PSNR and AMMBE values are shown in below table.

|  | RMSE | PSNR | AMMBE |
| --- | --- | --- | --- |
| Normal | 81.29± 0.01 | 10.44±0.89 | 0.09472±0.001 |
| Grade-I | 83.96± 0.07 | 9.72±0.06 | 0.094±0.05 |
| Grade-II | 99.81± 0.04 | 8.4±0.84 | 0.07±0.01 |
| Grade-III | 75.81± 0.2 | 10.6±0.4 | 0.13±0.05 |

DIC images contain the information of medium refractive index fluctuation information. The 3D ST amplitude spectra of stromal region of normal, grade-I, grade-II and grade-III cervical tissues are shown as Figure 4 (a-d), respectively. Mukhopadhyay et.al., already showed that for amplitude spectra analysis purpose, S-transform provides better results than wavelets. [9]

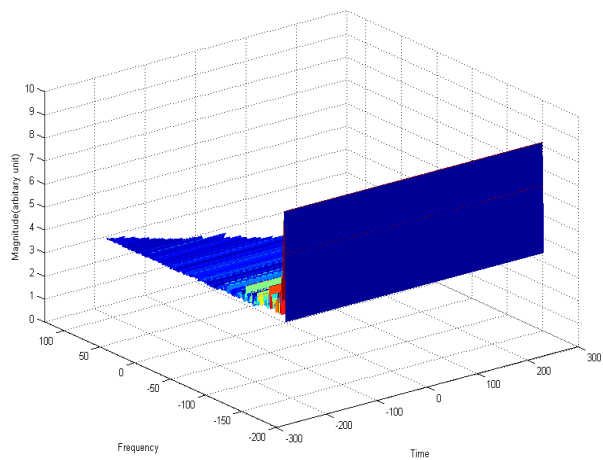

Normal
2(a)

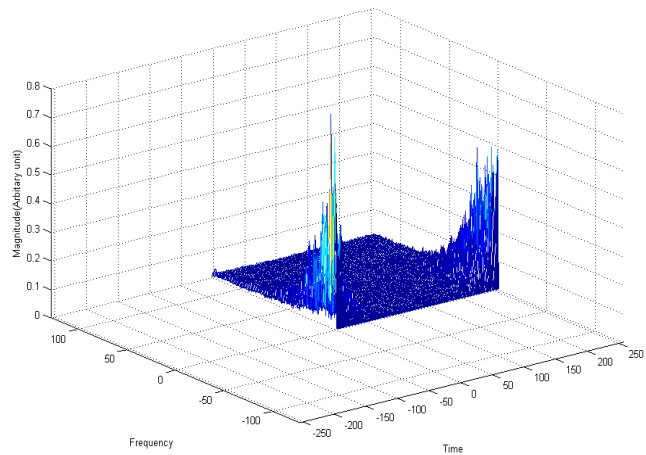

Grade-I
2(b)

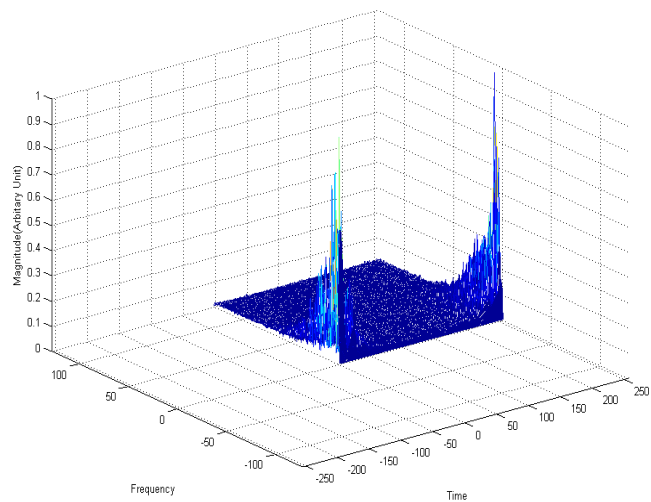

Grade-II

2(c)

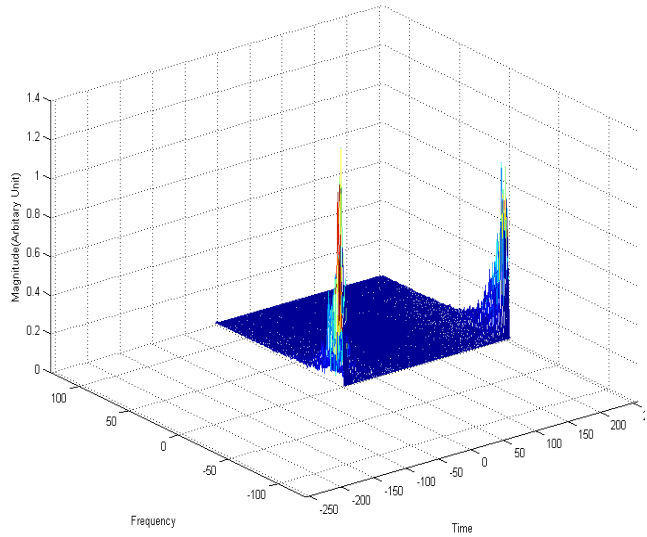

Grade-III
2(d)
Fig2: ST amplitude spectra of different grades of DIC images of Stromal region

The variations among 'peak of amplitude spectra' are shown in time-frequency domain in below table.

Peak of Amplitude Spectra in Time-frequency plane at the most dominant frequency in horizontal direction is shown in the below table.

| Normal | Grade-I | Grade-II | Grade-III |
|---|---|---|---|
| 7 ±0.01 | 0.5 ±0.009 | 0.9 ±0.007 | 0.955 ±0.008 |

Here, it is clearly visible that the peak of ST amplitude spectra has been decreased dramatically from normal to grade-I due to medium inhomogeneity. Thereafter we can observe the increasing trend of amplitude spectra with the progress of cancer among different grades.

## V. CONCLUSION

In conclusion, this investigation on sets of human normal, grade-I, grade-II and grade-III tissues sections using S-transform (ST) analysis of their DIC images shows potential to obtain diagnostic information from the spatial frequency distributions of these samples. With the evolution among disease tissues and the development from low grade to high grade disease tissues, the substantial changes with respect to normal tissue were observed in time-frequency domain. The variations occurred due to increment roughness of the medium among different grades of DIC tissues with normal DIC tissue. The HKMDHE based procedure has been performed on DIC images such that signal processing tools like S-transform can more prominently classify the normal and disease tissues. In a nutshell, this study clearly shows that it is possible to discriminate among the normal and different grades of disease tissues. These observations on ST of tissue refractive index variations may prove to be valuable for developing light scattering approaches for non-invasive diagnosis of pre- and early-stage cancer. Authors hope that this work will expand the use of ST based analysis into the field of biomedical optics for cancer research.

ACKNOWLEDGMENT

SM would like to thank the hospitality of IISER Kolkata where a part of this work was done.